*Original Article*

# Emotion Recognition for Challenged People Facial Appearance in Social using Neural Network

P. Deivendran[1], P. Suresh Babu[2], G. Malathi[3], K. Anbazhagan[4], R. Senthil Kumar[5]

[1]*Department of IT, Velammal Institute of Technology Panchetti, Chennai.*
[2]*Department of IT, Velammal College of Engineering and Technology, Madurai.*
[3]*School of Computer Science and Engineering, Vellore Institute of Technology, Chennai.*
[4]*Department of Computer Science and Engineering, Saveetha School of Engineering, SIMATS.*
[5]*Department of Information Technology, University of Technology and Applied Science, Muscat, Sultanate of Oman.*

[1]deivendran1973p@gmail.com

***Abstract*** *- Human communication is the vocal and non-verbal signal to communicate with others. Human expression is a significant biometric object in picture and record databases of surveillance systems. Face appreciation has a serious role in biometric methods and is good-looking for plentiful applications, including visual scrutiny and security. Facial expressions are a form of nonverbal communication; recognizing them helps improve the human-machine interaction. This paper proposes an idea for face and enlightenment invariant credit of facial expressions by the images. In order on, the person's face can be computed. Face expression is used in CNN (Convolutional Neural Network) classifier to categorize the acquired picture into different emotion categories. It's a deep, feed-forward artificial neural network. Outcome surpasses human presentation and shows poses alternate performance. Varying lighting conditions can influence the fitting process and reduce recognition precision. Results illustrate that dependable facial appearance credited with changing lighting conditions for separating reasonable facial terminology display emotions is an efficient representation of clean and assorted moving expressions. This process can also manage the proportions of dissimilar basic affecting expressions of those mixed jointly to produce sensible emotional facial expressions. Our system contains a pre-defined data set, which was residential by a statistics scientist and includes all pure and varied expressions. On average, a data set has achieved 92.4% exact validation of the expressions synthesized by our technique. These facial expressions are compared through the pre-defined data-position inside our system. If it recognizes the person in an abnormal condition, an alert will be passed to the nearby hospital/doctor seeing that a message.*

## 1. Introduction

The pasture of facial appearance has to make significant progress. Successful models enclosed be developed to relocate the facial look of emotion acted by the face image of the human object being [1]. Such a model regularly thinks that a broad range of facial expressions of the intention of someone is available [12]. Collecting this place may not always be feasible. Those models exist and need only one face unbiased fascia picture of the purpose of an object [8]. Such methods generally synthesize terms of essential emotion. However, in actual life, faces illustrate combined necessary emotions [20]. For instance, one can be happily or fearfully surprised. Therefore, we require a model that (a) can synthesize realistic expressions presenting together fundamental and assorted emotions and (b) require a single-face picture is taken as input. Here this paper deals with one model, therefore, introduced as an Expression Map. The XM was exposed to be functional in estimating the fraction of unlike essential emotion in a facial. For example, a facial look may show 35% happiness and 55% sadness [17]. The modern method utilizes the XM to notice the size of a particular grouping of basic affecting terminology on a known appearance balanced face image. The AMD facial appearance, as used in the living being histogram in usual humanity, is unsuitable for invention [15]. Designed for mixture, use a particular set of features from on behalf of facial shape, texture, and look makeup and look intensity. In a key variation between features and old in and here is so as toward in the past technique the feel characteristic is separated into low and haughty occurrence mechanism which is process independently [22]. This will helps us in dropping the sound in the synthesized facial. This facial appearance prepares the XM and subsequent learning method of Self Organizing Map (SOM). Imagine that every expression can modify in appearance commencing with the look of a nonaligned face. It might be changing and saved as a model in the node [11]. The reason for an objective function that chooses an exact model of a join the train XM in a given sensation and a face and impartial target facial emergence image [5]. To look on behalf of the merger of two essential emotions. Our progression takes a partial grouping of two methods represented through each node. Accordingly, the communicative face from end to end adds together. The grouping of patterns toward the object of the face image looks sensible with a conserved focus to identify the image.

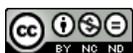 

## 2. Literature Survey

Each face is an ordinary signal used by a human and can be expressed depending upon their mood. A set of attempts to build a model in a facial expression analysis [2]. That has been requested in numerous fields like robotics, gaming, and medical facilities to help the system [10]. For this reason, in the twentieth century [11], Ekman has to define how different types of emotions are explained. Society as a human life with different types of expressions like irritation, fear, happiness, sad, dislike, disgust, and surprise. Here, an existing reading is going on facial appreciation and performance in a dataset to be established [6]. In current years, computers have more and more powerful computing control and huge data sets [7, 8]. Machine learning algorithms are compared to traditional methods [12]. The machine learning algorithm integrates two feature extraction processes [5] and classification [13]. The operation process can mechanically extract the internal facial appearance of the sample data [15], has dominant feature extraction capabilities, and is related to computer vision (CV). Computers can simply identify face expressions [14] along with determining personnel and including the amusement like social media, content-based system, fairness, and the healthcare system. Here are different approaches, such as wavelet and coefficients [1]. Zhang has explained in this paper that a lesser resolution (64x64) is enough [18]. Every human sensation is capable of the image segregated into different levels such as joy, unhappiness, repulsion, irritation, fright, and shock [19].

Meanwhile, the working mechanism is enhanced by combining the performance's image, voice, and textual data in a series of competitions. Mollahosseini has explained [22, 23] that the purpose of deep learning in the CNN algorithm can be an accessible database. Later than extracting the face from the data set, each image was reduced to (48x48) pixels [20]. Every structural design contains different layers and adds two origin styles in this method [17]. There is a convolution layer dimension consisting of 1x1, 3x3, and 5x5. It can be represented by the ability to use the network varies from system to system. While increasing the dimension of the image and the performance of the network layer is low. Those techniques are also possible to decrease the over-fitting problem [24]. In the crash of data processing, all types of networks are used to find the performance and face image classification technique [25]. The purpose of a new CNN algorithm is to detect AU faces in a network [9]. Here, two convolution layers are used to find the max-pooling in a connected layer. The numbers are used to activate and detect the parts of the image, which was explained by Yolcu [12]. After getting the image classification into the CNN algorithm, they can crop the period and find the key value. The iconic expression can be obtained from every image analyzed by employing the CNN algorithm to perceive face image appearance.

## 3. Facial Expression Sigmoid Function

The block diagram Fig.1 represents the essential and varied expressions synthesized through the frame separation approach. Observing the synthesized CNN training is taken as an input for feature extraction of the target image. Different steps have been taken for the alert message to pass to the hospital. The CNN classifier will accept the training data and check the different types of facial expressions by using the XM classifier.

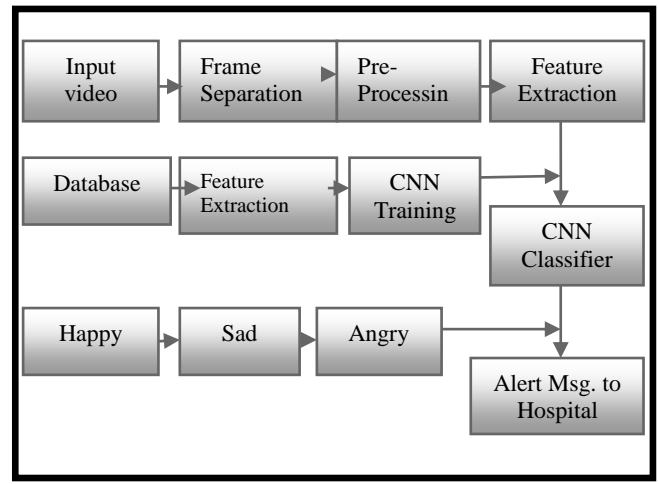

**Fig. 1 Architecture flow diagram**

The synthesized expressions can be appeared in wrinkles, furrows, and teeth and look ordinary on the face shape of the object. Here the output of F(x) is a function concerning Z2, and the beginning value is +1, and the end of the value is -1, which is an exponent of expression $e^x$ in Z2. It will appear in the calculation chart using frontward propagation and reverse propagation, and the result is simply a Sigmoid of Z2. Thus, $\partial O/O Z_2$ is efficiently derived from the function of Sigmoid(x).

$$F(x) = (1+e^{-x})^{-1}[1-(1+e^{-x})^{-1}]$$

$$F(x) = \text{sigmoid}(x)(1-\text{sigmoid}(x))$$

$$\partial O/\partial Z_2 = (O)(1-O)$$

In caring information [13], a convolution of the network is classified in a group of bottomless neural networks. Most of them regularly work to evaluate and illustrate all images [11]. It can be identified and shift to variant or gap invariant in networks. So that the inactive of the shared-weights plan and transformation of all types of network characteristics. It contains applications within any image classifications to be analyzed by the well-defined network topology.[9, 14], and the economic circumstance chain is summarized in Section VI. The future algorithm requires only one face-neutral picture of the object as an individual. Related workings will be presented in the subsequent section [6, 10]. Section III deals with the quality of the image partitioning method have been discussed. This method can be grouped and explained in section IV, and section V gives output results [29]. Figure. 2, shown below, represents the face image using the facial image classification compared with the existing mechanism in a similar part of the image classification analysis.

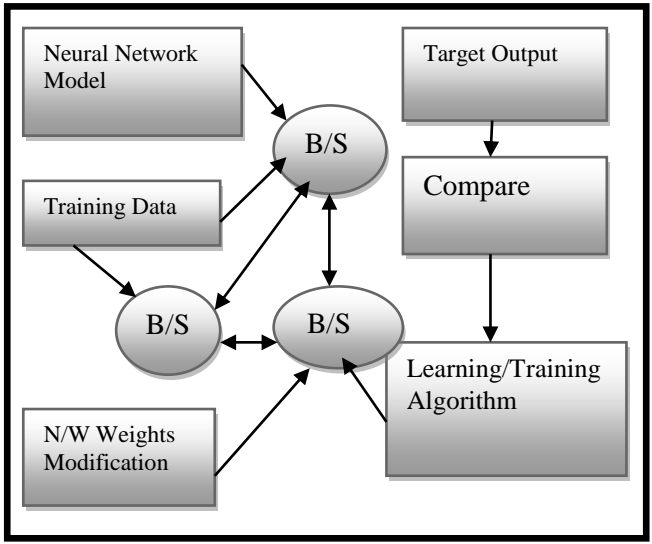

**Fig. 2 Functional diagram of a neural network model**

## 4. Classification Analysis & Probability

Individual steady space is an approach to face appreciation under uncontrolled conditions. Here usually exist many variations within face images taken under uncontrolled conditions, such as modifying their face, illumination, air, etc. Most of the previous plants are on face recognition, focus on exacting variations, and frequently assume the presence of others. This paper directly deals with face recognition below unrestrained conditions of the classifier[27]. The solution is the individual stable space (ISS), which only expresses private characteristics. A neural network name is planned for a rare face image keen on the ISS. Later on, three ISS-based algorithms are considered for FR below unrestrained conditions. There are no restrictions used for the images fed into these algorithms. In addition, to the different methods used, they do not need additional training to be tested [28]. These advantages construct them sensible to apply below-level unrestrained circumstances.

The existing algorithms are experienced on three huge face databases with a massive difference and understand greater performance than the existing FR techniques. This paper has explained a facade appreciation process that will appear at the top of PCA (Principal Component Analysis) and LDA (Linear Discriminated Analysis). The technique consists of two processes: initial, we plan the appearance picture as of the original vector space to an appearance subspace via PCA; succeeding us use LDA to attain a most excellent linear classifier. The fundamental design of combining PCA and LDA is to improve the simplification capacity of LDA when only a small number of samples per set are presented. Using PCA, we can build a face subspace during, which we apply, LDA to execute classification. The use of the FERET dataset can express a significant enhancement when primary components quite than unusual similes are fed in the direction of the LDA classifier.

$$\Pr(Y = k \mid X = x) = \frac{pi_k f_k(x)}{\sum_{l=1}^{K} pi_k f_k(x)}$$

Using the above formula reduced the dimension of the data points and image classification. However, the predictable data can be used to construct a partition of the image using Bayes' theorem. Let us assume that the value range is denoted by X. Let $X = (x_1, x_2, \ldots x_p)$ be derived from a multivariate Gaussian distribution. Here K is the number of data modules, Let Y is the response variable in $Pi_k$ is given an observation, and it is associated in $K^{th}$ class. The value $\Pr(X=x|y=k)$ is the number of possible functions. Let $f_k(x)$ be the big value if there is an elevated probability of an observation sample in the $K^{th}$ position of the variable X=x. The cross classifier with PCA and LDA provides a useful framework for other image recognition tasks.

### 4.1. Personalizing conservative composition Recommendation

Though a fan of traditional music was established to be below represented on top of social media and song stream platforms, they represent a significant target for the music recommender system. So we focus on this cluster of viewers and examine a large array of suggestion approaches and variants for the job of song artiste commendation. Inside the grouping of traditional music viewers, promote the assessment categorize users according to demographics and sequential music utilization manners. We describe the outcome of the beginning suggestion experiment and insight gained on behalf of the listener set less than thought.

## 4.2. Music personalized Reference System Based on Hybrid Filtration

Due to the tune's range and fuzziness and the music melody's correctness, the recommendation algorithm employing peak accuracy cannot completely match the user's analysis. For such difficulty, this paper proposes a cross-reference algorithm based on the joint filter algorithm and harmony geneses and designs an adapted music proposal system [27]. The scheme's first computer suggestion consequences according to the shared filter algorithm and realizes the potential benefit to the customer. Then every music is biased by liking on top of the genes of composing music. Later than load selection, the song with earlier preference is taken as a suggestion [25]. Lastly, two suggested outcomes were performed, weighted grouping and filter to make commendation. The investigational data point out the enhanced method can raise the correctness of recommendations and meet users' demands from different levels.

## 5. Implementation

### 5.1. Input Video
The live video taken from the camera is taken as the input video.

### 5.2. Frame Separation
Surround processing is the first step in the environment subtraction algorithm. This step schemes to classify the customized video frames by removing noise and unnecessary items in the frame to increase the quantity of information gained from the frame. Preprocessing an image is a method of collecting easy image processing tasks that modify the uncooked input video information into a system. It can be processed by following steps. Preprocessing the record is essential to improve the finding of touching objects, For example, by spatial and earthly smoothing, snow as disturbing plants on a tree.

### 5.3. Image pre-processing
- Image Representation is mainly classified into the following terms.
- Import the image using acquisition tools;
- Analyzing and testing the image;
- Output can be reported, which is based on analyzing that image.

### 5.4. Elimination
Feature mining is a type of dimensionality decrease that proficiently represents as an image compact vector technique. This approach is useful when large image sizes are reduced based on the required tasks such as illustration, matching, and retrieval.

### 5.5. Database
The database contains a pre-defined face pattern from feature pulling out with which the user's face is compared and emotion is detected.

### 5.6. CNN Algorithm
**Step-1:** frame = camera. read()

**Step-2** if (frame = imutils.resize(frame, width=500))
Assign new frame=gray

**Step-3** Detection of face
Faces=face_detection.detectMultiscale(gray, scalefactor=1,0, Minneighbors=12,minsize=(60,60),flags=cv2.casecadescale_image)

**Step-4** If(canvas=np_zeros((500, 700, 3),type="uint12"))
 then assign frame=newframe
frameClone = frame.copy()

**Step-5**
If(len(faces)>0))
Set the value 0, 1; faces = sorted(faces, reverse=True,

**Step-6** compares the number of key value in arry
Key1=lambda_x,(x[3]-x[0])*(x[1]-x[0])
If(Fx,Fy,Fh)=vces

**Step-7** To get the output of image color from grays_cale image, and resize to be fixed size in (28x28 )pixels

**Step-8**
To assign the ROI values for each classification via the CNN
Roi=gray[fY:fY+fH,fX:fX+fW]
Roi=img_aray(roi)

**Step-9** To get the dimension of the image size
Roi=roi.type("float") / 255.0

**Step-10** Find the ROI and probability
Roi=np.expand_dims(roi,axis==0)
Pre=emotion_classifier.predict(roi)[0,1]

**Step-11**
Label1=emotions["angry","disgust","scared","happy", "sad", "surprised", "neutral"]
if label=='happy':
VarHappy=VarHappy+1

**Step-12** check the type of emotion
if label=='sad':
VarSad=VarSad+1
If(VarSad)>Thresh:
if label=='angry':
Varangry=Varangry+1
ifVarangry>Thresh:

**Step13** To check the classification
 if label=='surprised':
Varsurprised=Varsurprised+1
ifVarsurprised>Thresh:
if label=='disgust':
Vardisgust=Vardisgust+1
If Vardisgust>Thresh:

*5.7. Classification*

Artificial neural networks are used in various classification work like image and audio. Different types of neural networks are used, from predicting the series of images to using regular neural networks. In particular, an LSTM, in the same way for image classification, uses of convolution neural network. This algorithm will intellect the face of emotions and send the mail to the consumer when irregular facial emotions are found.

*5.8. Alert Message*

Simple mail transfer protocol is a type of request layer protocol. When the client requests to send the mail to TCP protocol and link to SMTP protocol, the server can send the mail through the link-layer protocol. The SMTP server monitors every node based on that protocol rules. The network layer listens for a TCP relation from any client. The SMTP protocol to initiate a connection is established to the particular port number. After establishing the TCP protocol, the client sends the mail immediately to the requester side. The simple mail transfer protocol has been used to send the mail to a specific user.

# 6. Result and Analysis

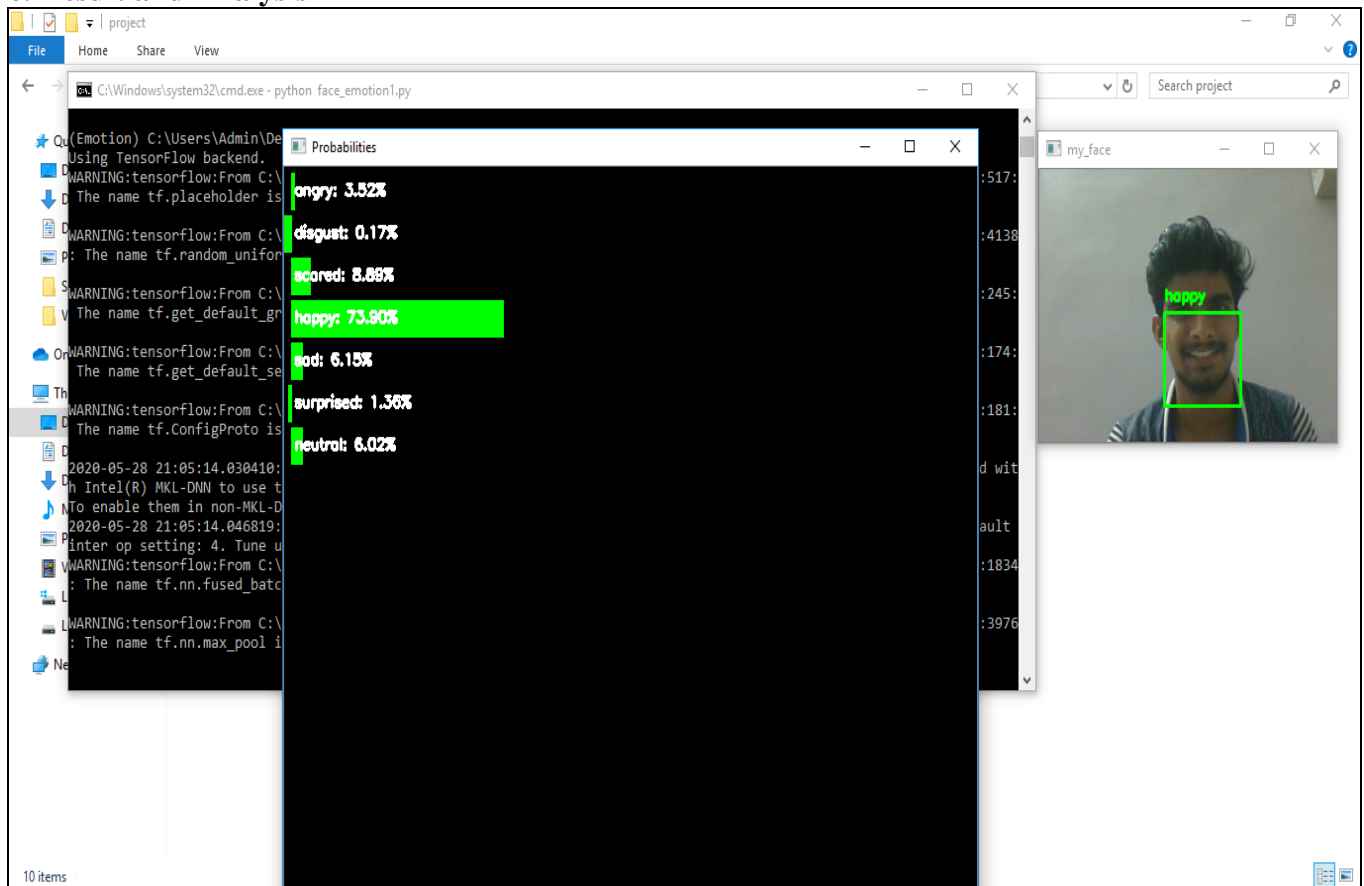

**Fig. 3 Neutral face**

The above Fig.3 is a neutral face of the result, here angry=0.82%, diggust=0.15%, scared=7.89, happy=22.18%, sad=8.10%, surprised=1.33%, neutral = 53.85%, so the neutral value is higher than the other attributes.

The graph Fig.4 shows the performance and comparison using the facial classifier technique, here tressed=2.6, sleepy=2.68, tired=3.08, walking=2.36, wake up 3.24, coordination =2.224 and fall as sleep =2.24, so the final output of the tired is the maximum percentage.

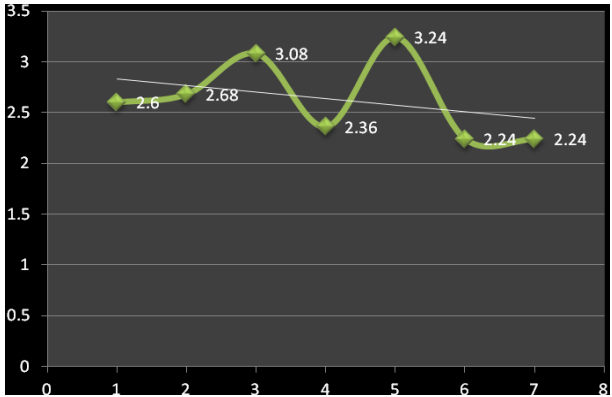

Fig. 4 Performance Comparison facial classifier

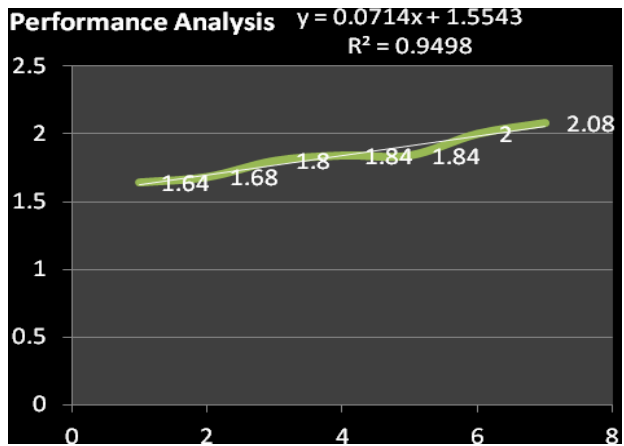

Fig. 5 Performance analysis of face recognition

The above graph, Fig.5, shows the performance and comparison using frame separation methods. Here the value falls as sleep is 2.08 and the $R^2$ value is 0.9498; together, the classifier techniques and the frame separation method have been analyzed. Finally, the facial classifier output of the image is more accurate and efficient.

## 7. Conclusion

This method is used to find the strange between facial images to classification images. The irregular facial images are detected by computing another part of segmentation. So the result can be obtained from video images and compared to the dataset collected from the present method. It is completed and defined by using the convolution neural networks concept. This effort is an idea to make it easy for the patients as well as community people who are living alone. Make use of the network layer to classify the facial expressions identical to the pre-defined data set and validate the patient's condition.